\documentclass[runningheads]{llncs}
\usepackage{graphicx}
\usepackage{comment}
\usepackage{amsmath,amssymb} % define this before the line numbering.
\usepackage{color}
\usepackage{subfigure}
\usepackage{array}
\usepackage{tabularx}
\usepackage{float}
\usepackage{epsfig}
\usepackage{multirow}
\usepackage[width=122mm,left=12mm,paperwidth=146mm,height=193mm,top=12mm,paperheight=217mm]{geometry}

\begin{document}
	% \linenumbers
	%\pagestyle{headings}
	%\mainmatter
	%\def\ECCVSubNumber{2848}  % Insert your submission number here
	
	\title{Local Context Attention for Salient Object Segmentation} % Replace with your title
	
	%\begin{comment}
	%\titlerunning{ECCV-20 submission ID \ECCVSubNumber} 
	%\authorrunning{ECCV-20 submission ID \ECCVSubNumber} 
	\author{Jing Tan, Pengfei Xiong, Yuwen He, Kuntao Xiao, Zhengyi Lv}
	\institute{Megvii Research \\
	    \email{\{tanjing, xiongpengfei, heyuwen, xiaokuntao, lvzhengti\}@megvii.com}
	    }
    %\end{comment}
%******************

%\begin{comment}
%\author{First Author\inst{1}\orcidID{0000-1111-2222-3333} \and
%Second Author\inst{2,3}\orcidID{1111-2222-3333-4444} \and
%Third Author\inst{3}\orcidID{2222--3333-4444-5555}}
%\institute{Princeton University, Princeton NJ 08544, USA \and
%Springer Heidelberg, Tiergartenstr. 17, 69121 Heidelberg, Germany
%\end{comment}
%
%******************
\maketitle

\begin{abstract}
% saliency seg需要将不同类别的显著性物体与背景区分开来。尽管显著性物体没有语义一致性，但具有很强的位置信息。基于这个先验，本文提出了一种新的local context的方法来做显著性物体分割。
Salient object segmentation aims at distinguishing various salient objects from backgrounds. Despite the lack of semantic consistency, salient objects often have obvious texture and location characteristics in local area. Based on this priori, we propose a novel Local Context Attention Network (LCANet) to generate locally reinforcement feature maps in a uniform representational architecture. The proposed network introduces an Attentional Correlation Filter (ACF) module to generate explicit local attention by calculating the correlation feature map between coarse prediction and global context. Then it is expanded to a Local Context Block(LCB). Furthermore, an one-stage coarse-to-fine structure is implemented based on LCB to adaptively enhance the local context description ability. Comprehensive experiments are conducted on several salient object segmentation datasets, demonstrating the superior performance of the proposed LCANet against the state-of-the-art methods, especially with \textbf{0.883 max F-score} and \textbf{0.034 MAE} on DUTS-TE dataset.

\end{abstract}

% ############################################introduction########################################
\section{Introduction}

Salient object segmentation aims at locating the most obvious and salient objects from a given image. It has been widely used in various challenging fields like automatic focus, autonomous driving, scene understanding, image editing, etc. In the past decades, salient object segmentation approaches \cite{PoolNet,AFNet,MLMSNet,Amulet,DCL,DSS,CapsNet,PAGE-Net,HRSOD,CNet,CapSal,EGNet} have already obtained promising performances on various benchmarks \cite{ECSSD,DUTS,DUT-OMRON,HKU-IS,PASCAL-S}. Nevertheless, most of the previous salient object segmentation methods treat it as a general semantic segmentation problem, which improves the performance by increasing the semantic receptive field, optimizing the edge accuracy or other methods. 

\begin{figure}
\begin{center}
   \includegraphics[width=1.0\linewidth]{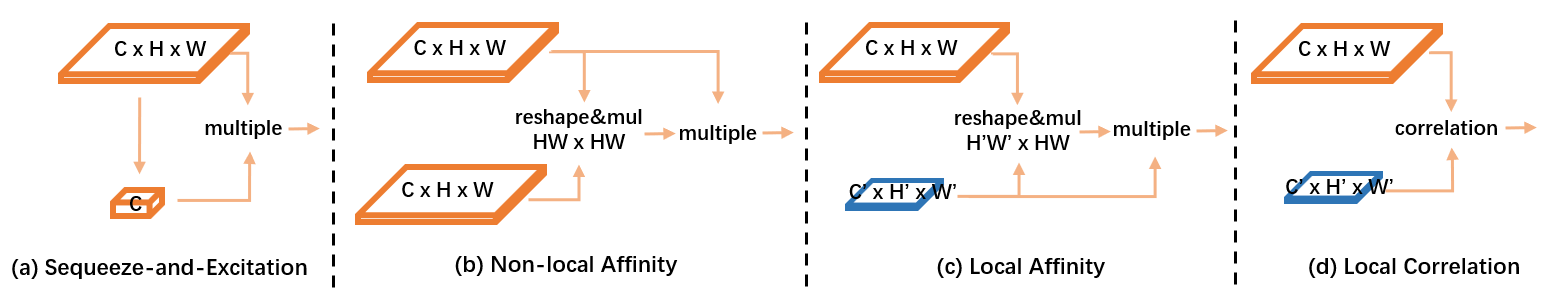} 
\end{center}
   \caption{Comparison of different attention approaches. From left to right: (a) Squeeze-and-Excitation\cite{senet}. (b) Non-local Affinity\cite{nonlocal}. (c) Local Affinity and (d) Local Correlation Attention. It can be noticed that, Local Correlation is operated on the local context of global feature, instead of the entire feature map.}
\label{fig:attentions}
\end{figure}

%Their essence is to make foreground objects more distinguishable. However, this implicit approach often leads to other problems. Liu et. al.\cite{PoolNet} proposed a method that can utilize the in-network multi-scale features to enhance the feature description capability of different layers， which is also crucial to increasing receptive fields. Although the global and local context are preserved simultaneously, region features of objects are hidden in the global feature map thus feature consistency are sacrificed across different scales. This leads to a distinct decrease in the small objects, may even damage the accuracy of large objects.
%

As a pixel-level classification problem, the intrinsic properties of salient object segmentation determine that it is different from traditional semantic segmentation. Salient objects usually do not belong to the same category, and their textures and shapes are various, which makes it hard to distinguish the salient objects and the background by simply increasing the receptive field. 

In contrast, salient objects often have obvious local context characteristic of the image. Despite their various sizes and locations, their vivid texture are always quite different from the surrounding backgrounds. Based on this prior, we 
believe that the calculation of each pixel attention on an equal basis is computationally inefficient and out of focus in the previous approaches. Therefore, we rethink salient object segmentation task from a more macroscopic point of view, that separating the salient object from the distant irrelevant background, and then extracting local context features related to the object as supervision to enhance the distinguish ability of foreground object.

%and improve the accuracy of salient segmentation by extracting local context features related to the object as a guide to the importance, which is attention.

An intuitive way to extract local context features is to construct a coarse-to-fine architecture with multi-scale inputs\cite{HKU-IS,Multi-scale}. It generates an approximate prediction in the first stage, then crops the image as the input of a second refine network. Nevertheless, this method relies heavily on the accuracy of coarse predictions and does not make good use of the relationship between global and local contexts due to the cropping operation in the coarse stage. In addition, the increase of inference time makes the coarse-to-fine methods less favorable for practical applications. A global scene of the image can provide global semantic information, while local context around the target object produces the relationship between foreground and background. Both of them provides useful hints for inferring the content of the target. Therefore, how to strengthen local context features while retaining global context features becomes the key to improve salient object segmentation accuracy. To this end, we propose a novel Local Context Attention Network (LCANet) to adapt global and local features with a uniform representational power.

%
%Both the coarse-to-fine and the feature pyramid methods share the same motivation that models should have different receptive fields for objects of different scales. When applied on small objects, local context is more important than global context. Despite the inefficiency, the coarse-to-fine methods sacrife the feature consistency across multi-crop images. The fine results depend on the coarse results, which means it still need a enough accurate network. In contrast, feature pyramid method fully utilizes the representational power of the model, but transform objects of all scales equally. This leads to a distinct decreasehttps://www.overleaf.com/project/5e3a63ac7c767d00018ccda2 on the small objects. 

%The task of saliency segmentation demands inherently consideration in both local region context and global scene context. 

Specifically, the proposed LCANet is mainly built upon a one-stage coarse-to-fine architecture. A coarse segmentation network is built from a standard classification model to extract the discriminant feature, which is up-scaled to generate the coarse feature map. Then an Attentional Correlation Filter (ACF) module is designed to generate local context attention. The local feature map is cropped based on the coarse location after image processing, and is regarded as the correlation filter to make a convolution with the whole feature map. With the help of convolution, the correlation feature map is taken as an attention map to concatenate with the original feature to explicitly enhance the local context description ability. In contrast to other attention modules \cite{senet,nonlocal,ocnet} exploring the channel or spatial weight, the proposed ACF module retains the global receptive field with local context enhanced in the surrounding areas, as shown in Figure \ref{fig:framework} (d). Furthermore, ACF module is enhanced to Local Context Block (LCB) by Multi-scale ACF operator and a Local Coordinate Convolution (LCC) layer. The LCC layer adopts the relative coordinates of the coarse prediction as another additional feature map to adaptively incorporate the local region context into the global scene context in the spatial dimension. Based on the enhanced LCB, an one-stage coarse-to-fine network is constructed in a type of encoder-decoder architecture as shown in Figure~\ref{fig:framework}. A multi-stage decoder is designed to aggregate the high-level information extracted by coarse network to gradually refine the segmentation results. The LCB is also implemented in the stages of decoder to handle with various sizes of salient objects.

In summary, there are three contributions in our paper:

\begin{itemize}
%\item We present a new investigation results about the influence of %the scale variation on salient object segmentation with quantitatively %and qualitatively analysis. To our best
%knowledge, we are the first to conduct experiments to explore scale %variation in this field.

\item We rethink the salient object segmentation task from the intrinsic properties of salient object, and create a newly model structure, Local Context Attention Network (LCANet), to strengthen local context features by adaptively integrating the local region context and global scene context in one stage coarse-to-fine architecture.

\item We design a novel Local Context Block (LCB), on the basis of Attentional Correlation Filter (ACF). As a basic module, it can be used in many situations, instead of the traditional global and non-local attention map.

\item Detailed experiments on five widely-used benchmarks indicate the effectiveness of our proposed modules and architecture in improving the accuracy. We achieve state-of-the-art performance on all of these datasets with thorough ablation studies, especially a new record of \textbf{0.883 max F-score} and \textbf{0.034 MAE} without any other refinement steps on DUTS-TE dataset. 
\end{itemize}

% ############################################# related works ########################################
\section{Related Work}
\textbf{Salient Object Segmentation:}
In the past decades, a number of approaches for saliency detection are developed. Most of them regard salient object segmentation as a special case of semantic segmentation, try to increase receptive field through multi-level features\cite{DSS,RFCN,Amulet,BDMPM,ELD,dsr}. They think high-level features in deep layers encode the semantic information of an abstract description of objects, while low-level features keep spatial details for reconstructing the object boundaries. Some researchers think edge information is the key to improve segmentation accuracy. \cite{RFCN,DCL} adopt post-processing heuristics to obtain the refine predictions. DEA\cite{DEA} simply uses an extra loss to emphasize the detection error for the pixels within the salient object boundaries, while others \cite{Contour,Objectcontour} consider semantic contour information from a pretrained contour detector. Although these methods are proved to be effective, no one analyzes the problem from the perspective of salient objects. Local context is easily overlooked in salient object segmentation. 

\textbf{Attention Module:}
Attention module is one of the most popular operations in recent neural networks to mimic the visual attention mechanism in the human visual system. SE-Net\cite{senet} explores a channel-wise attention map and has achieved state-of-the-art performance in image classification. In the field of semantic segmentation, several methods \cite{pspnet,deeplabv3+,pan} adopt multi-scale attention map to increase the receptive field of the high-level features. EncNet\cite{encnet,attentionseg} introduces context encoding to enhance the prediction that is conditional on the encoded semantics. Non-Local\cite{nonlocal,ocnet} is further proposed self-attention with non-local affinity matrix for vision tasks. By contract, we apply the local context attention map on the output of encoder network to integrate the local region and global scene description.

\textbf{Correlation Filter:}
Correlation filter \cite{KCF,MOSSE,CSK} has proved to be effective in most of single object tracking methods. It takes advantage of the fact that they can specify the desired output of a linear classifier for several translations and image shifts based on the dot-product at different relative translations of two patches using only a fraction of the computational power. It is relevant to local similarity between the object and its local neighborhood and suitable for convolution operator. Naturally, encoding the local correlation feature as an attention map is an immediate thought. 

\textbf{Coarse-to-fine:}
These also exist researchers finding the optimal coarse-to-fine solutions to deal with the scale-variation problem. However, most of them focus on the feature expression. DSS \cite{DSS} proposes a saliency method by introducing short connections to the skip-layer structures within a multi-layer HED architecture. NLDF\cite{NLDF} combines the local and global information through a multi-resolution grid structure. Amulet\cite{Amulet} directly aggregates multi-level features by concatenating feature maps from different layers. Although the features from deeper layers help to locate the target, objects with different sizes cannot be represented in the same feature structure. A more proper way is to employ the multi-scale features in a coarse-to-fine fashion and gradually predict the final saliency map. However, the increase of inference time makes the coarse-to-fine methods less favorable for practical.

% ######################################LCA Network###########################################
\section{Methodology}
In this section, we first detailedly introduce the idea of the Attentional Correlation Filter (ACF) module, and elaborate how this module specifically handles the local context issue. After that, we describe the complete Local Context Attention Network (LCANet) based on the Local Context Block (LCB), which builds a coarse-to-fine structure in one stage encoder-decoder architecture.

\begin{figure*}
\begin{center}
\includegraphics[width=1.0\linewidth]{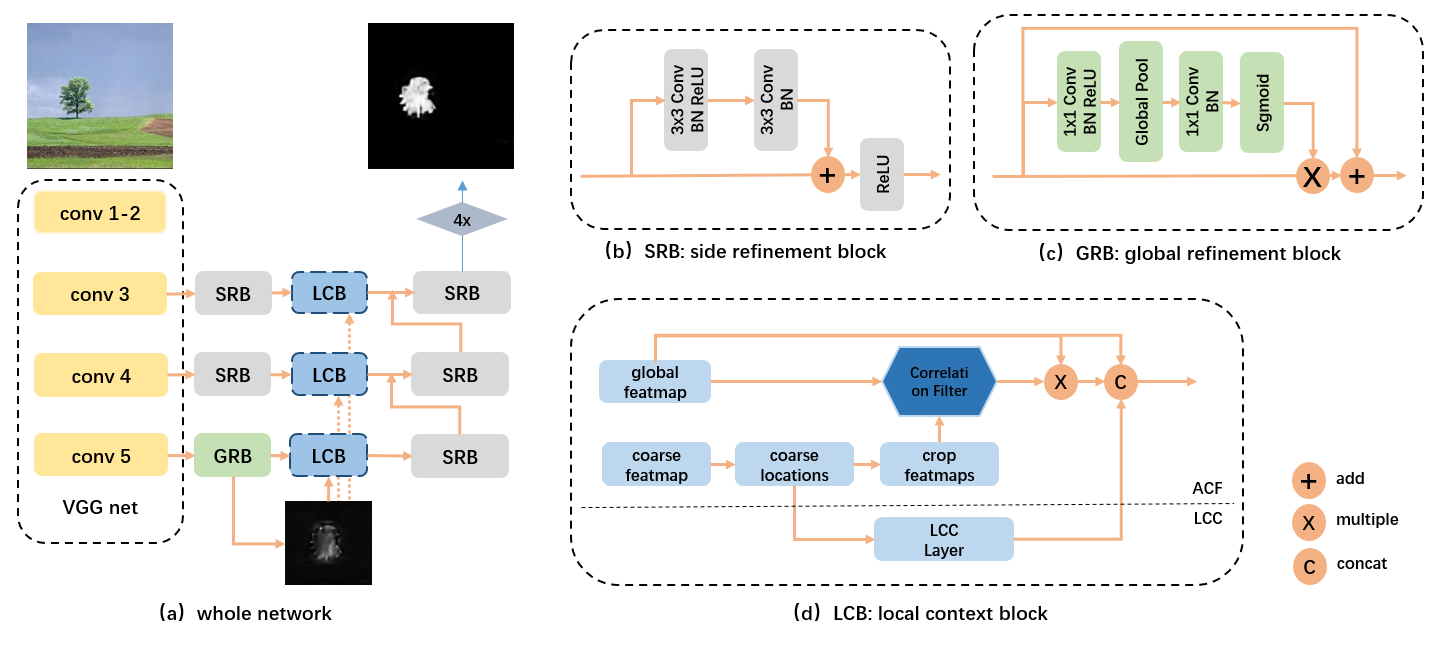}
\end{center}
   \caption{Overview of our Local Context Attention Network (LCANet). A coarse network is built on a VGG network followed by GRB to generate coarse prediction of the input image. Then the feature map is entered into LCB to produce local context attention maps of the input global features. LCB consists of ACF and LCC modules. It can be implemented on multi-stage of decoder network, cooperating with SRB to gradually refine the final segmentation results.}
\label{fig:framework}
\end{figure*}

\subsection{Attentional Correlation Filter}\label{sec:ACF}
%While the general two-stage coarse-to-fine method relies heavily on the precision of the coarse model, we adopt the idea of attention map to adaptively incorporate local context and global feature description in a uniform representational architecture. Inspired by the correlation filter in single object tracking, we design an Attentional Correlation Filter module to calculate the correlation map between coarse prediction and high-level discriminant feature.
%下面这段之后改掉
%There is a natural prior to saliency segmentation, which is to focus on strong features in a given image, perhaps bright colors or the center position, etc. Based on this prior, we think that it is a waste of computation and missing the point to consider each pixel equally when calculating the attention map. A better method is to find these significant features and directly calculate the correlation between other pixel and the most significant features, and get the correct attention map through the local context.

Let's revisit the main prior of salient object segmentation. Saliency object usually has unique feature representations that are different from the surrounding background, such as vivid colors or clear boundaries. Pixels of salient object has strong correlation with its local context in physical space. Based on this observation and analysis, we propose an Attentional Correlation Filter(ACF) module to strengthen the salient features, which apply the attention mechanism to enhance feature expression ability of local context in global feature.

\textbf{Correlation Filter.} Given a target object $T\in \mathbb{R}^{H' \times W'} $ and an reference image $I\in \mathbb{R}^{H \times W}$. For each pixel $(x,y) \in I$ , we have the correlation map $Corr$ between $I$ and $T$ is calculated as 

\begin{equation}
Corr(I,T)_{x,y} = \sum_{v=-\frac{W'}{2}}^{\frac{W'}{2}}\sum_{u=-\frac{H'}{2}}^{\frac{H'}{2}} T[u,v] \odot I[x+u, y+v]
%Corr(I,T) =  I[L(p)] \odot T ,\  for\  p\  in\ H \times W 
\label{equ:coorlation}
\end{equation}

The correlation map can be generated according to the sliding window on the reference image. In the field of single object tracking\cite{CSK,KCF,MOSSE}, it is used widely to obtain the similarity degree between the tracking object and its neighborhood scene to search for the update object position in the next frame. The higher the correlation value, the higher the probability that two patches belong to the same object. The operation of the sliding window is similar to convolution. We regard the feature of the target object as convolution weight, then the correlation feature map can be generated by convolution on the original global feature.

\begin{figure}
\begin{center}
\includegraphics[width=1.0\linewidth]{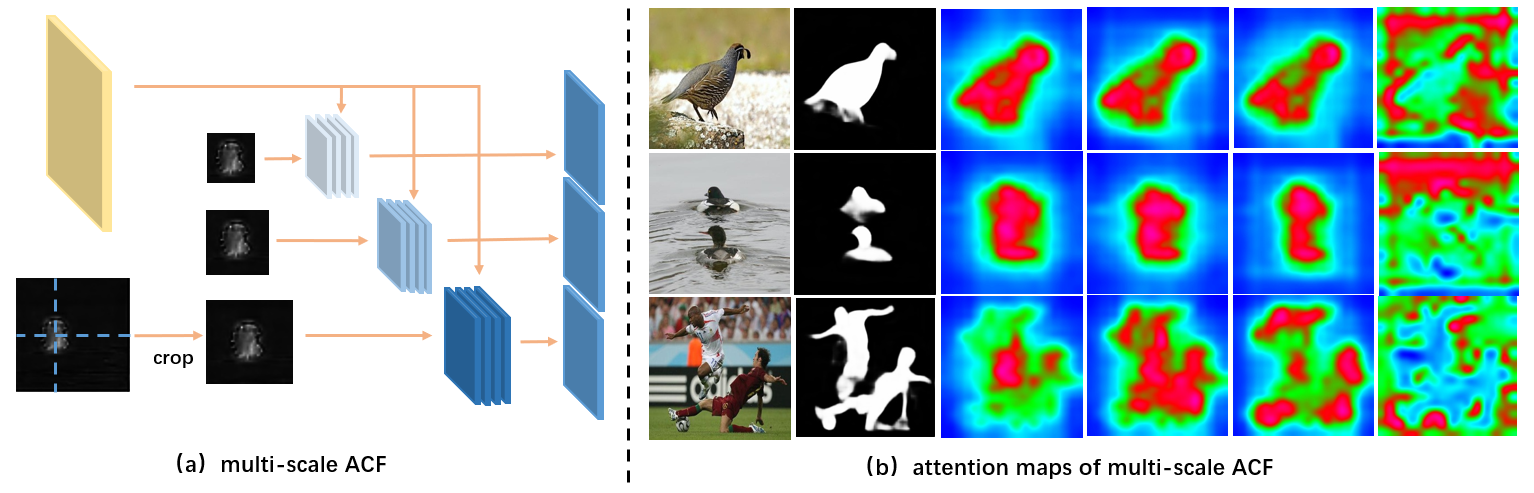}
\end{center}
   \caption{Multi-scale ACF. In (a), multi-scale local feature maps are cropped based on coarse prediction and convoluted with the original feature to generate attention maps. In (b), attention maps of ACF with various scales (line 3,4,5) from coarse results (line 2) are compared with the attention map of non-local module (line 6).}
\label{fig:ACF}
\end{figure}

\textbf{Attentional Correlation Filter.} We use the generated correlation feature map as an attention layer for subsequent processing. Given a local feature map $T\in \mathbb{R}^{C'\times H'W'} $ and global feature map $I\in \mathbb{R}^{C \times HW}$, ACF can be generated as, 

\begin{align}
%Att_{ij}=\frac{exp(Att_{ij})}{\sum_{c=1}^{C}exp(Att_{ij})}\;
%Att_{acf}=Sigmoid(I \otimes T)
ACF = I \odot Sigmoid(Corr(I,T))
\end{align} 

There are many ways to implement attention map, as compared in Figure~\ref{fig:attentions}. In global attention \cite{senet} or non-local affinity attention \cite{nonlocal,ocnet}, features are used to calculate the pixel similarity in channel or spatial dimensions. When it acts on the local area, the similar local affinity map can be generated by multiplying the local and global features. However, it is unreasonable that the correlation in channel dimensions loses the spatial relationship between pixels. The direct feature fusion will lead to the migration between features. In ACF, we crop the high-level feature from encoder into local feature map based on the coarse prediction. The coarse object location is calculated from the maximum and minimum positions of binarized course predictions. Then the coarse feature map is warped into a given size based on the affine transformation. The generated local context feature map is normalized to keep the same channel of global feature map and odd kernel size, and convoluted with the original feature to generate attention feature map. As shown in Figure~\ref{fig:ACF}, non-local attention map puts more emphasis on bright areas than significant areas, while ACF significantly enhances the feature discrimination ability in the region of interest. 

\subsection{Local Context Block}\label{sec:LCB}
We further extend the proposed ACF module to a more robust local context descriptor. Firstly, we introduce a Local Coordinate Convolution (LCC) layer to increase the local context information in the spatial perspective. Secondly, the ACF is extended to multi-scale ACF to better cope with the significant changes of different sizes of objects, and inaccurate results of coarse network. 

\textbf{Local Coordinate Convolution.} 
%除了现在的特征层面的local context，一个空间上的local context isimplemented here.
The calculation of Local Coordinate Convolution Layer $lcc_x, lcc_y$ of target pixel $p_t=I(x_t,y_t)\in I$ can be formulated as following:

\begin{equation}
lcc_x(x,y) = [ 1 - \frac{|x - x_t|}{H}] ,\ (x,y) \in H \times W
\end{equation}
\begin{equation}
lcc_y(x,y) = [ 1 - \frac{|y - y_t|}{W}] ,\ (x,y) \in H \times W
\end{equation}

we concatenate the LCC and $I$ in channel dimension, and then send them into the convolution layer. The standard coordinate convolution is developed to provide an explicit relative position relationship on the spatial feature map. Based on the coarse location of the salient object, we modify it to further enhance local features in the perspective of physical space. $lcc_x$ and $lcc_y$ are respectively calculated through the distance of pixel relative to the center point of coarse prediction. The nearer the distance, the greater the value. LCC module, combined with ACF module, extracts the local context information of coarse results in the dimensions of feature and space dimensions, in order to complement each other in local area enhancement.

\textbf{Multi-scale ACF.} Furthermore, in order to compensate the error of the coarse prediction, we apply the multi-scale attention maps. Based on the rough segmentation result, the original coarse feature map is warped into different scales. We extract local characteristics with scale changes by adjusting the expansion of the external rectangle of coarse predictions. All these multi-scale attention maps are 
respectively multipled and concatenated with the global features and to generate the local enhanced feature maps. As shown in Figure~\ref{fig:ACF}, with the implementation of Multi-scale ACF, the significance of objects with different sizes is highlighted on different attention maps. On the contrary, non-local attention more strengthens the distribution of global features, while ignoring the salient objects.

\subsection{Network Architecture}
In the task of semantic segmentation, it has been proved that high-level features and multi-scale decoder are the main factors to improve the segmentation results. While the performance of LCB depends on the result of coarse network, the first stage of LCANet needs to obtain acceptable coarse prediction results under controllable computational complexity. Then a multi-stage decoder is needed to further refine the coarse results. Based on the above knowledge, we propose two complementary modules that are are capable of capturing the positions of salient objects and refining their details from coarse level to fine level. 

\textbf{Coarse Network.} 
The most direct way to integrate coarse and refined results is sharing the encoder layers. High-level features generated from the encoder network already have preliminary classification ability in terms of distinguishing salient objects and backgrounds. We applied a standard VGG\cite{VGG} network as the backbone network to extract multi-layer discriminating feature maps. Following the high-level semantic features from VGG, a Global Refinement Block (GRB) is applied to improve receptive field and change the weights of the features to enhance the semantic consistency. As illustrated in Figure~\ref{fig:framework}, two 1x1 convolutions and a global pooling operations are applied to reallocate the feature map and generate a spatial attention map onto the high-level features. This global attention map explicitly makes feature maps be aware of the locations of the salient objects. Then coarse prediction is generated by up-sampling this global strengthened features. 

\textbf{Coarse-to-fine Network.} \label{sec:baseline} 
Based on the coarse network and LCB, we construct an one-stage coarse-to-fine network. We build our network based on a type of classic U-shape architecture as shown in Figure~\ref{fig:framework}. Although GRB improves the receptive filed to capture the global information of the input images, the classification abilities in different stages of backbone network are ignored resulting in diverse consistency manifestation. So we design a residual structure consisting of two 3 x 3 convolution operations to iteratively fuse discriminant features of different scales, which is named as side refinement block (SRB). Different from previous symmetrical encoder-decoder architectures, we only take the result of stage3 quadrupled as the final output. The last two abandoned up-sampling convolution blocks save a lot of computation, but improve the accuracy in our experiments. As mentioned above, the LCB is implemented cooperating with SRB to strengthen local context features of each stage before up-sampling. Similar to multi scale operations in feature level, multi-stage of LCB is also essential to adapt to the wrong predictions and scale variations as much as possible.

\textbf{Loss Function.} The LCANet is trained in an end-to-end manner with losses of coarse and refine outputs. we use the cross-entropy loss between the final saliency map and the ground truth. Besides that, we apply a boundary preservation loss to enhance the edge accuracy. Sobel operator is adopted to generate the edge of ground truth and predication saliency map of network. Then the same cross-entropy loss is used to supervise the salient object boundaries. The similar boundary supervision is used and proved in several previous works\cite{PoolNet,PFA,BDMPM}. Different from them, Sobel operator makes the model pay more attention to the junction of salient objects and background. The total loss function of the LCANet is the weighted sum of the above losses. 

\begin{equation}
L = \lambda_0 L_{cs} + \lambda_1 L_{rf} + \lambda_2 L_{cs_{bd}} + \lambda_3 L_{rf_{bd}}
\end{equation}

Where $L_{cs}$ and $L_{rf}$ denote the cross entropy loss function of coarse and coarse-to-fine predictions, and $L_{cs_{bd}}$ and $L_{rf_{bd}}$ are their boundary preservation loss. An experimental weight are applied to combined them all. Also, online hard example mining\cite{OHEM} strategy is adopted with $L_{cs}$ and $L_{rf}$ respectively during training. 

In contrast to other two-stage coarse-to-fine networks, LCANet is more like inserting a layer of local context attention into a standard encoder-decoder network. It doesn't separate the coarse and fine networks with multi times inference. What's more, it adaptively combines the local surrounding context and global scene information in a uniform feature description.

% ######################################Experiments###########################################
\section{Experiments}
In this section, we mainly inverstigate the effectiveness of the proposed LCB module. While the performance of LCB depends on the result of coarse network, we first compare the effect of each module in the complete network structure. Then we conduct more experiments on the configurations of LCB to analyze the influence of LCB on local context in detail. Finally, we compare the proposed LCANet with other state-of-the-art approaches quantitatively and qualitatively. 

\subsection{Experimental Setup}
\textbf{Datasets.} To evaluate the performance of the proposed approach, we conduct experiments on five popular benchmark datasets\cite{DUTS,ECSSD,DUT-OMRON,HKU-IS,PASCAL-S}. These datasets all contain a large number of images as well as well-segmented annotations and have been widely used in the filed of salient object segmentation. DUTS\cite{DUTS} is the largest dataset containing 10,553 images for training and 5,019 images for testing. Both training and test sets in DUTS contain very complex scenarios with high content variety. ECSSD\cite{ECSSD} contains 1,000 natural images manually selected from the Internet. HKU-IS\cite{HKU-IS} includes 4447 images with multiple disconnected salient objects overlapping the image boundary. DUT-OMRON\cite{DUT-OMRON} has 5,168 images with many semantically meaningful and challenging structures. Images of this dataset have one or more salient objects and relatively complex background. PASCAL-S\cite{PASCAL-S} contains 850 natural images that are free-viewed by 8 subjects in eye-tracking tests for salient object annotation. 

\textbf{Evaluation Metrics.} In order to obtain a fair comparison with other state-of-the-art salient object segmentation  approaches, we train the proposed networks on DUTS training set (DUTS-TR), and evaluate them on DUTS test set(DUTS-TE) and the other four datasets. For quantitative evaluation, two universally-agreed, standard metrics, mean absolute error ($MAE$) and maximum F-measure ($maxF$) are adopted respectively \cite{DSS}. F-measure reflects the overall performance of pixel classification. It is computed by weighted harmonic mean of the precision and recall. $MAE$ indicates the average pixel-wise absolute difference between the estimated saliency map and ground-truth.

\textbf{Implementation Details.} All the networks mentioned below follow the same training strategy. A VGG-16 pre-trained on Imagenet\cite{imagenet} is used to initialize the convolution layers in the backbone network. 
The parameters in other convolution layers are randomly initialized. All training and test images are resized to $256\times 256$ before being fed into the network. They are trained using mini-batch stochastic gradient descent (SGD) with batch size 48, momentum 0.9, weight decay $1e-5$ and 300 epochs. As the common configuration, the "poly" learning rate policy is adopted where the initial rate is multiplied by $(1-\frac{iter}{max\_iter})^{power}$ with power 0.9 and the base learning rate is set as $1e-4$. Data augmentation contains mean subtraction, random horizontal flip, random resizing with scale ranges in [0.8, 1.2], and random cropping to keep most of the salient object intact for training. 

\begin{figure}
\begin{center}
\includegraphics[width=1.0\linewidth]{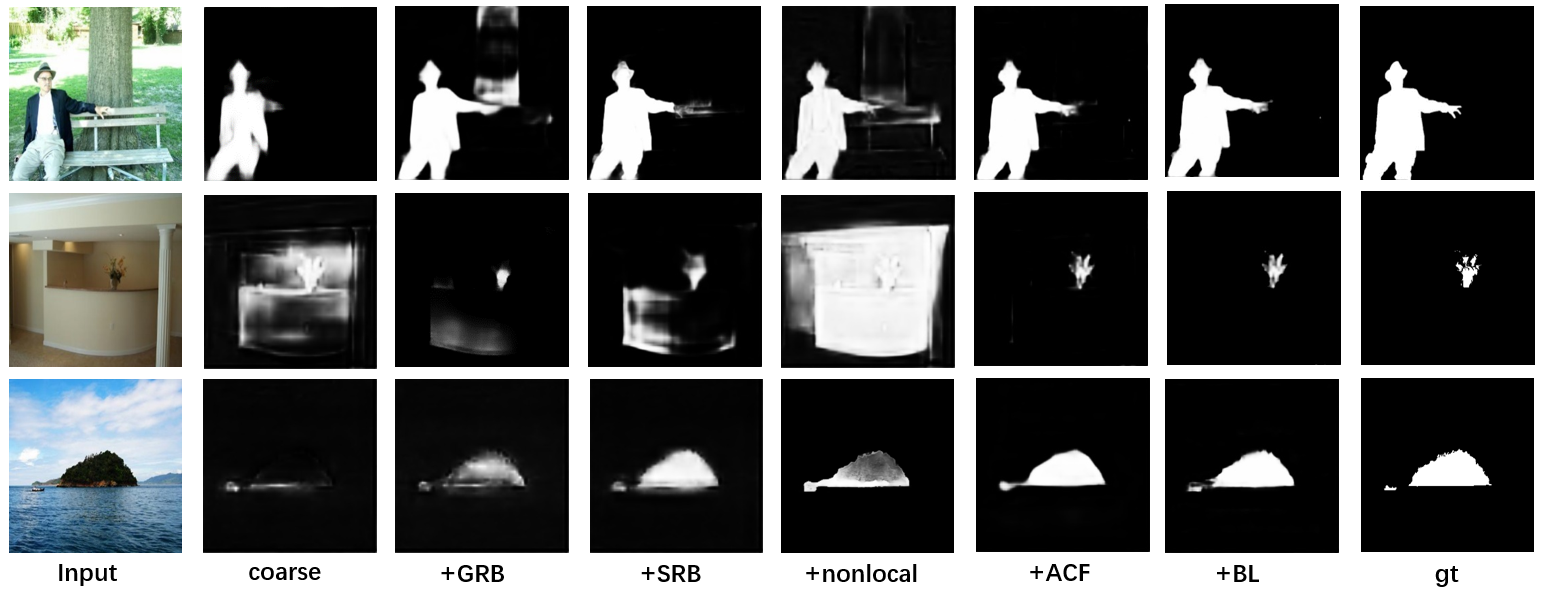}
\end{center}
   \caption{Result comparisons of different modules in our proposed method. From left to right, each line depicts results of coarse backbone model, and the successive additions of GRB, SRB decoder, non-local attention map, ACF, boundary preservation loss (BL) and the ground truth.}
\label{fig:samples}
\end{figure}

\subsection{Ablation Studies of Main Architecture}

The main architecture is an encoder-decoder structure consisting of GRB, SRB and LCB. Besides that, boundary loss (BL) is verified here as part of main network. We conduct ablation experiments on two challenging datasets DUTS-TE and ECSSD to demonstrate the effectiveness of each module separately. 

\textbf{GRB.} Firstly, we verify the accuracy of coarse output. According to Table~\ref{tab:mainablation}, the addition of GRB gives performance gains in term of both $maxF$ and $MAE$ on the two datasets over the coarse network. While implemented on the coarse network, the $maxF$ of DUTS-TE is improved from 0.775 to 0.818. Also the performance is improved further to 0.843 while applied onto the coarse-to-fine network. The corresponding visual comparisons can be found in Figure \ref{fig:samples}. The global pooling operator helps to increase the receptive field of the network and highlight the areas of interest. 

\textbf{SRB.} Here SRB refers to the decoder network containing the SRB module. As can be observed as in Table~\ref{tab:mainablation}, simply embedding of SRB decoder helps improve the performance on both $maxF$ and $MAE$. When GRB and SRB are superposed concurrently, the performance of the model is further enhanced, which indicates the effectiveness of multi-stage refinement for solving the aliasing effect of up-sampling. 

\begin{table}
\caption{Ablation analysis for the main architecture on two popular datasets. All experiments are trained on the union set of DUTS-TR. No.1 depicts the encoder network, and the other modules are successively embedded. As can be observed, each proposed module plays an important role and contributes to the performance.}
\begin{center}
\setlength{\tabcolsep}{2.2mm}
\label{ablation}
\begin{tabular}{c|cccc|cc|cc}
\hline
\multirow{2}{*}{No.} &\multirow{2}{*}{GRB} & \multirow{2}{*}{SRB} & \multirow{2}{*}{ACF} & \multirow{2}{*}{BL} & \multicolumn{2}{c|}{DUTS-TE} & \multicolumn{2}{c}{ECSSD} \\
\cline{6-9}
& & & & &$maxF$$\uparrow$ & $MAE$$\downarrow$ & $maxF$$\uparrow$ & $MAE$$\downarrow$ \\
\hline
\hline
1 & & & & & 0.775 & 0.081 & 0.863 & 0.080  \\
2 & \checkmark & & & & 0.818 & 0.051 & 0.883 & 0.066  \\
3 & & \checkmark & & & 0.830 & 0.047 & 0.891 & 0.058  \\
4 & \checkmark & \checkmark & & & \textbf{0.843} & \textbf{0.046} & \textbf{0.915} & \textbf{0.054}  \\
\hline
5 & & & \checkmark & & 0.781 & 0.077 & 0.871 & 0.078  \\
6 & \checkmark & & \checkmark & & 0.847 & 0.044 & 0.904 & 0.051  \\
7 & \checkmark & \checkmark & \checkmark & & 0.873 & 0.039 & 0.928 & 0.034  \\
8 & \checkmark & \checkmark & \checkmark & \checkmark & \textbf{0.875} & \textbf{0.038} & \textbf{0.931} & \textbf{0.033}  \\
\hline
\end{tabular}
\end{center}
\label{tab:mainablation}
\end{table}

\textbf{ACF.} Furthermore, we implement ACF onto the above encoder-decoder network. As the core of LCB module, ACF can well verify the effectiveness of local context attention. Here we only apply the simplest single-scale ACF onto the high-level features generated from coarse network. When it directly acts on the coarse prediction, the accuracy is only slightly improved. This proves from the side that a more reasonable local context is the key to improve the accuracy. When the coarse result is poor, ACF is similar to the attention map on the global context. When it is added to GRB and SRB in turn, the role of ACF is highlighted, and both $maxF$ and $MAE$ on the two datasets are further improved. As shown in Figure \ref{fig:samples}, the local guidance information generated by ACF allows our network to focus more on the salient objects, and greatly improve the quality of resulting saliency maps.

\textbf{BL.} Finally, the boundary preservation loss (BL) is adopted to further improve the quality of boundary in the produced saliency map. Although the performance is only slightly improved, the upper edge of the visual results was optimized to be closer to the ground truth in Figure \ref{fig:samples}. 

\subsection{Ablation Studies of LCB}

To demonstrate the effectiveness of our proposed LCB, we adopt two basic networks. $coarse$ depicts the GRB enhanced coarse network, and $baseline$ is the combined model of GRB, SRB and BL, which has been verified in the previous section. We study the different variants of ACF and LCC modules in LCB, and compare them with other attention methods. 

\textbf{Coarse-to-fine Structures.} By introducing local context into the baseline network, the most direct way is to execute the model twice by cropping the predictions of the first network as the input of the second one. In Table~\ref{tab:lcbablation}, we conducted the experiments of executing the $coarse$ and $baseline$ twice respectively. The performance of $baseline\times2$ is significantly improved both in terms of $maxF$ and $MAE$, while the accuracy of $coarse\times2$ is slightly decreased. This is mainly because the refine network is heavily dependent on the results of coarse network. However, when $baseline + ACF$ integrates the local context attention and the global feature map in a joint training manner, the performance is greatly improved as shown in Table~\ref{tab:lcbablation}. 

%\begin{table}[H]
\begin{table}
\caption{Ablation analysis for the proposed main architecture on DUTS-TE. Different types of ACF and LCC modules are verified and compared with other attention modules. As can be noticed, ACF is superior to others, and LCC also contributes to the performance.}
\begin{minipage}{0.45\linewidth}
\centering
\begin{tabular}{l|c c}
\hline
Model & $maxF$$\uparrow$ & $MAE$$\downarrow$  \\
\hline\hline
coarse $\times$ 2 & 0.813 & 0.050 \\
baseline $\times$ 2 & 0.861 & 0.042 \\
\hline
baseline + $Att_{se}$ & 0.845 & 0.047 \\
baseline + $Att_{lf}$ & 0.847 & 0.046 \\
baseline + $Att_{nl}$ & 0.861 & 0.041 \\
\hline
baseline + ACF & \textbf{0.875} & \textbf{0.038}  \\
\hline
\end{tabular}
\end{minipage}
\begin{minipage}{0.55\linewidth}  
\centering
\begin{tabular}{l|c c}
\hline
Model &$maxF$$\uparrow$ &$MAE$$\downarrow$  \\
\hline\hline
baseline + $ACF_{0.1}$    & 0.874 & 0.038 \\
baseline + $ACF_{0.3}$    & 0.871 & 0.041 \\
baseline + $ACF_{0.5}$    & 0.875 & 0.038 \\
baseline + $ACF_{msz}$    & 0.879 & 0.037 \\
\hline
baseline + $ACF_{msz}$ + LCC  & 0.882 & 0.041 \\
baseline + $ACF_{msz+msa}$ + LCC & \textbf{0.883} & \textbf{0.034} \\
\hline
\end{tabular}
\end{minipage}

\label{tab:lcbablation}
\end{table}

\textbf{Other Attention Modules.} Also we compare the proposed ACF module with other attention-based modules. $Att_{se}$, $Att_{nl}$ and $Att_{lf}$ are effective modules of squeeze-and-excitiation\cite{senet}, non-local affinity\cite{nonlocal}, and a modified local affinity depicted in Figure \ref{fig:attentions}. As can be observed from Table~\ref{tab:lcbablation}, although the accuracy can be improved by increasing the receptive field, these modules are still worse than the proposed ACF module. $Att_{lf}$ is similar to ACF by generating the affinity matrix bewteen local context and global feature map. Nevertheless, the performance is even worse than results of $baseline \times2$, we think that the multiplication of affinity matrix in different feature space destroys the ability of feature description of both global and local feature map.

\textbf{Multi-scale ACF.} Furthermore, we investigates the effectiveness of Multi-scale ACF. Figure \ref{fig:ACF} depicts the visualization results of attention maps produces by the cropped feature maps with different sizes. Also the performance of Multi-scale ACF are showed in Table~\ref{tab:lcbablation}. It can be observed both in qualitative or quantitative analysis that different attention map corresponds to different size of salient object. And Also the unknown segmentation errors of coarse network also affect the performance of ACF. Although the accuracy of ACF with different scales fluctuates, Multi-scale ACF can cope with these accidental changes, and further improve the performance of both $maxF$ and $MAE$. 

\textbf{LCC.} LCC is another complementary module in LCB. As present in Table~\ref{tab:lcbablation}, LCC further improves the result of Multi-scale ACF from 0.879 to 0.882. This proves the complementary relationship between ACF and LCC. Under the guidance of an explicit location provided by LCC, it realizes a more accurate and robust segmentation on the final result. The final LCB consists of Multi-scale ACF and LCC. It can be implemented onto the different stages of decoder to further enhance the final performance as $ACF_{msz+msa}$ depicted in Table~\ref{tab:lcbablation}.

\subsection{Comparisons with the State-of-the-Arts}

We compare the proposed LCANet with sixteen recent state-of-the-art methods on five datasets including PoolNet\cite{PoolNet}, PFA\cite{PFA}, AFNet\cite{AFNet}, MLMSNet\cite{MLMSNet}, CPD\cite{CPD}, BDMPM\cite{BDMPM}, GRL\cite{GRL}, PAGRN\cite{PAGRN}, Amulet\cite{Amulet}, SRM\cite{SRM}, UCF\cite{UCF}, DCL\cite{DCL}, DHS\cite{DHS}, DSS\cite{DSS}, ELD\cite{ELD}, NLDF\cite{NLDF}. We obtain the saliency maps of these methods from authors or the deployment codes provided by authors for fair comparison.

\begin{table*}
\footnotesize 
\caption{Quantitative comparisons of the proposed approach and sixteen state-of-the-art CNN based salient object detection approaches on five datasets. The best two scores are shown in {\color{red}red} and {\color{blue}blue}.}
\begin{center}
\begin{tabular}{c c c c c c c c c c c}
\hline
%\multirow{}{}{}
%\multicolumn{1}{c}{datasets} &
& 
\multicolumn{2}{c}{DUTS-TE} & 
\multicolumn{2}{c}{ECSSD} &
\multicolumn{2}{c}{HKU-IS} &
\multicolumn{2}{c}{PASCAL-S} & 
\multicolumn{2}{c}{DUT-OM} 
 \\
\cline{2-11}
&$maxF$ &$MAE$ &$maxF$ &$MAE$ &$maxF$ &$MAE$ &$maxF$ &$MAE$ &$maxF$ &$MAE$ \\
\hline\hline
%RFCN\cite{RFCN}           &0.782&0.089 &0.896&0.097 &0.886&0.080 &0.855&0.115 &0.738&0.094 \\
ELD\cite{ELD}             &0.737&0.092 &0.867&0.081 &0.840&0.073 &0.788&0.122 &0.719&0.090 \\
DHS\cite{DHS}             &0.811&0.065 &0.904&0.062 &0.890&0.053 &0.845&0.096 &-&-  \\
DCL\cite{DCL}             &0.785&0.081 &0.895&0.079 &0.889&0.063 &0.845&0.111 &0.756&0.086 \\
UCF\cite{UCF}             &0.772&0.111 &0.901&0.070 &0.887&0.062 &0.849&0.109 &0.729&0.120 \\
SRM\cite{SRM}              &0.826&0.058 &0.915&0.056 &0.905&0.046 &0.867&0.085 &0.769&0.069 \\
Amulet\cite{Amulet}        &0.777&0.084 &0.913&0.060 &0.896&0.051 &0.861&0.098 &0.742&0.097 \\
NLDF\cite{NLDF}           &0.812&0.064 &0.903&0.065 &0.901&0.048 &0.851&0.100 &0.753&0.079 \\
DSS\cite{DSS}             &0.813&0.064 &0.895&0.064 &0.901&0.047 &0.850&0.099 &0.760&0.075 \\
PAGRN\cite{PAGRN}          &0.854&0.054 &0.923&0.064 &0.917&0.047 &0.869&0.094 &0.770&0.070 \\
GRL\cite{GRL}              &0.834&0.050 &0.923&0.044 &0.913&0.037 &0.881&0.079 &0.778&0.063 \\
BDMPM\cite{BDMPM}          &0.851&0.048 &0.925&0.048 &0.920&0.039 &0.880&0.078 &0.774&0.064 \\
CPD\cite{CPD}              &0.864&0.043 &0.936&0.040 &0.924&0.033 &0.866&0.074 &0.794&0.057 \\
MLMSNet\cite{MLMSNet}      &0.851&0.049 &0.928&0.045 &0.921&0.039 &0.862&0.074 &0.774&0.064 \\
AFNet\cite{AFNet}          &0.862&0.046 &0.935&0.042 &0.923&0.036 &0.868&0.071 &0.797&0.057 \\
PFA\cite{PFA}             &0.870&{\color{blue}\textbf{0.040}} &0.931&{\color{blue}\textbf{0.032}} &0.926&{\color{blue}\textbf{0.032}} &{\color{red}\textbf{0.892}}&{\color{blue}\textbf{0.067}} &{\color{red}\textbf{0.855}}&{\color{blue}\textbf{0.041}} \\
PoolNet\cite{PoolNet}     &{\color{blue}\textbf{0.880}}&0.041 &{\color{blue}\textbf{0.937}}&0.044 &{\color{blue}\textbf{0.931}}&0.033 &0.865&0.072 &0.821&0.056 \\
\hline\hline
\textbf{LCANet}                         &{\color{red}\textbf{0.883}}&{\color{red}\textbf{0.034}} &{\color{red}\textbf{0.939}}&{\color{red}\textbf{0.029}} &{\color{red}\textbf{0.931}}&{\color{red}\textbf{0.030}} &{\color{blue}\textbf{0.889}}&{\color{red}\textbf{0.064}} &{\color{blue}\textbf{0.843}}&{\color{red}\textbf{0.037}} \\
\hline
\end{tabular}
\end{center}
\label{tab:Quantitative}
\end{table*}

\textbf{Quantitative Evaluation.} Table~\ref{tab:Quantitative} shows the quantitative evaluation results of the proposed method and other state-of-the-art salient segmentation approaches in terms of $maxF$ and $MAE$. As present, our method outperforms other approaches across all these datasets. To be specific, LCANet achieves large improvement compared with the best existing approach on DUT-TE dataset. Both of $maxF$ and $MAE$ are definitely increased from the ever best PoolNet\cite{PoolNet} based on the VGG backbone. On PASCAL-S and DUT-OMRON, although the performance of $maxF$ is slightly lower than PFA\cite{PFA}, the $MAE$ exceeds it. We find that objects in them are large or multi-subjects. When the object size is large, local context is almost the same as global context the, the role of LCANet will be correspondingly weakened. While the proposed approach is a simple one-stage structure, it can be further improved with multi-branch learning algorithms.

\begin{figure*}
	\begin{center}
		\includegraphics[width=1.0\textwidth]{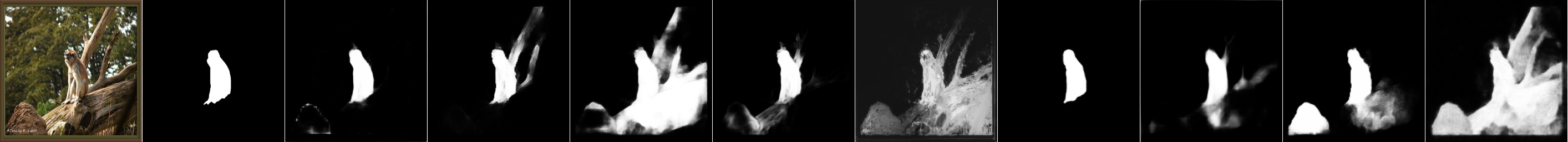}
		\includegraphics[width=1.0\textwidth]{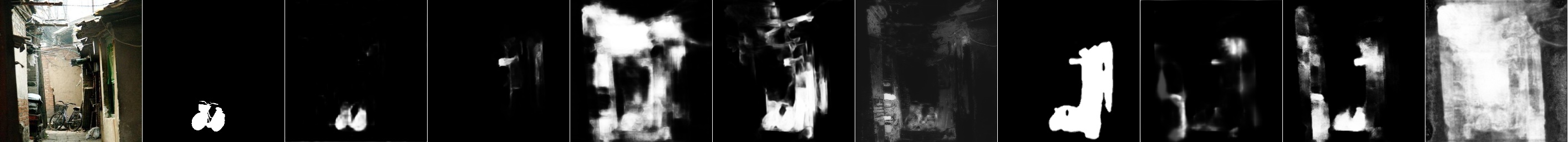}
		\includegraphics[width=1.0\textwidth]{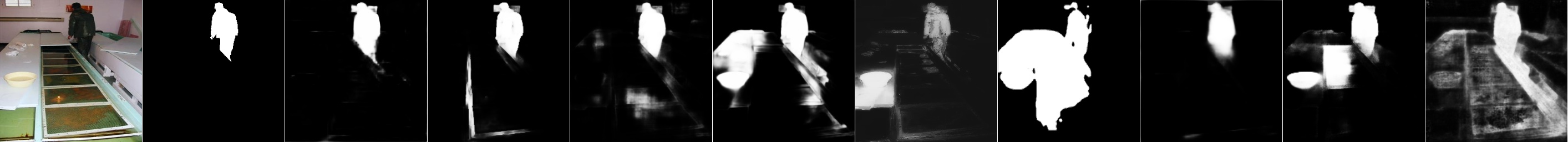}
		\includegraphics[width=1.0\textwidth]{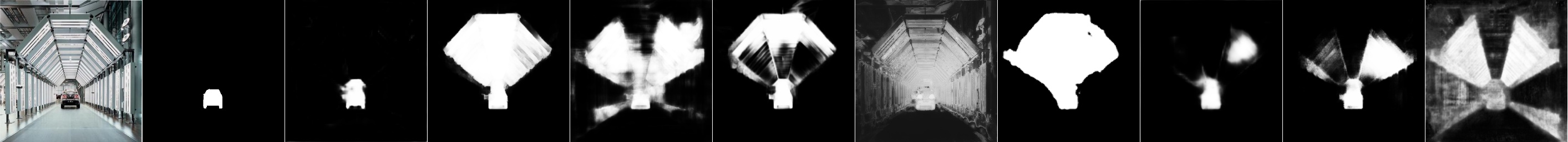}
		\includegraphics[width=1.0\textwidth]{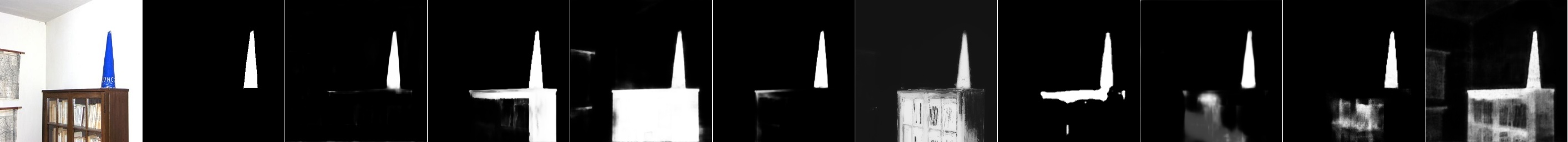}
		\includegraphics[width=1.0\textwidth]{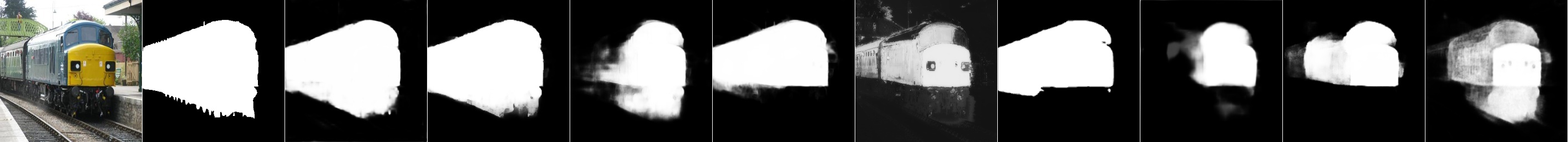}
		\includegraphics[width=1.0\textwidth]{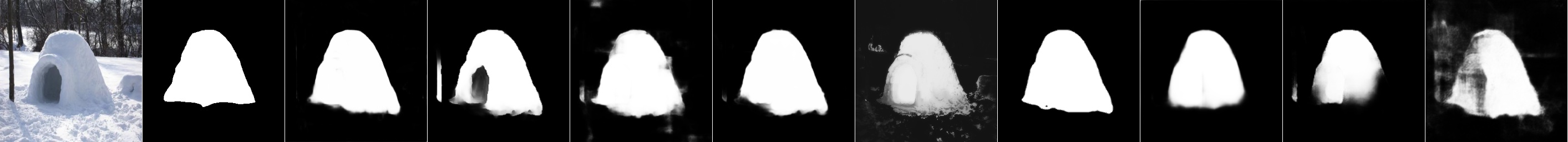}
		\includegraphics[width=1.0\textwidth]{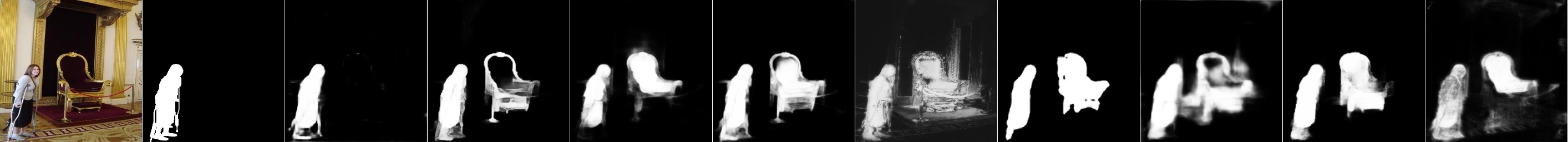}
        \includegraphics[width=1.0\textwidth]{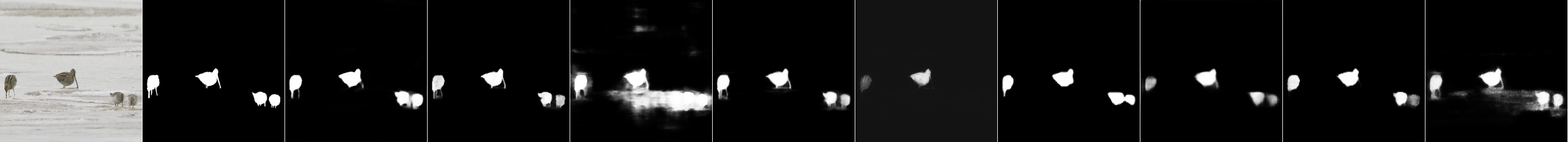}
		\includegraphics[width=1.0\textwidth]{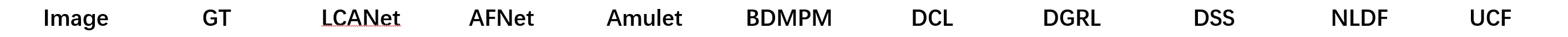}
	\end{center}
	\caption{Qualitative comparisons of the state-of-the-art algorithms and our proposed \textbf{LCANet}. GT means ground-truth masks of salient objects. }
	\label{fig:Qualitative}
\end{figure*}

\textbf{Qualitative Evaluation.} To further explain the advantages of our approach, Figure \ref{fig:Qualitative} provides a visual comparison of our method and other state-of-the-arts. From the former 5 rows of Figure \ref{fig:Qualitative}, it is clear that our method is obviously superior to others coping with small objects. While other methods are difficult to distinguish salient objects and background, LCANet obtains accurate segmentation results based on the coarse-to-fine guidance. This further verifies the effectiveness of local context in salient object location. More than this, we notice that the consistency of larger objects is also preserved as shown in row 6,7 of Figure \ref{fig:Qualitative}. Even when salient objects are scattered in different places of the image, there is still a certain probability not be affected. Also it still obtains a good segmentation prediction when the salient object located in the side of the image. These observations indicate the intergration of local and global context information is important to deal with salient object segmentation, regardless of the position and size of the object.
%This is attributed to the improvement of dynamic fusion of local region of interest based on the local context attention.
%In LCANet, the global receptive field is not destroyed, but only adjusted by the weight of the local context of interest. 

% #############################################Conclusion works########################################
\section{Conclusion}
In this paper, we propose a local context attention network to cope with salient object segmentation. Based on the prior that saliency object usually has unique feature representations that are different from the surrounding background, we proposed a Local Context Block consisting of an Attentional Correlation Filter and a Local Coordinate Convolution layer, in order to intergrate the local context information into the global scene features in one-stage coarse-to-fine architecture. Detailed experiments verify the feasibility of the proposed LCANet. It achieves comparable performance with other state-of-the-art methods based on a simple baseline. We believe that this model will be useful for other scenarios.

\clearpage
\bibliographystyle{splncs04}
\bibliography{egbib}
\end{document}